# Populating cellular metamaterials on the extrema of attainable elasticity through neuroevolution

Maohua Yan, Ruicheng Wang, Ke Liu*

*Department of Advanced Manufacturing and Robotics, Peking University, Beijing, China*

**Abstract:** The trade-offs between different mechanical properties of materials pose fundamental challenges in engineering material design, such as balancing stiffness versus toughness, weight versus energy-absorbing capacity, and among the various elastic coefficients. Although gradient-based topology optimization approaches have been effective in finding specific designs and properties, they are not efficient tools for surveying the vast design space of metamaterials, and thus unable to reveal the attainable bound of interdependent material properties. Other common methods, such as parametric design or data-driven approaches, are limited by either the lack of diversity in geometry or the difficulty to extrapolate from known data, respectively. In this work, we formulate the simultaneous exploration of multiple competing material properties as a multi-objective optimization (MOO) problem and employ a neuroevolution algorithm to efficiently solve it. The Compositional Pattern-Producing Networks (CPPNs) is used as the generative model for unit cell designs, which provide very compact yet lossless encoding of geometry. A modified Neuroevolution of Augmenting Topologies (NEAT) algorithm is employed to evolve the CPPNs such that they create metamaterial designs on the Pareto front of the MOO problem, revealing empirical bounds of different combinations of elastic properties. Looking ahead, our method serves as a universal framework for the computational discovery of diverse metamaterials across a range of fields, including robotics, biomedicine, thermal engineering, and photonics.

## 1. Introduction

Metamaterials are engineered materials whose extraordinary properties originate from the geometry of their microstructures rather than from the composition of their constituent materials [1–5]. Mechanical metamaterials, in particular, exhibit mechanical properties unattainable by material composition alone, necessitating the rational design of their internal architecture [6–19]. Typically, these materials are engineered using periodic unit cells that determine their overall properties. However, designing such complex unit cells has traditionally relied heavily on expert knowledge and extensive trial-and-error, prompting a shift towards automated design methodologies [20–28]. Metamaterial design is an ill-defined inverse problem with an infinite-dimensional geometrical design space and a one-to-many mapping from properties to microstructures [29–34]. Common approaches include topology optimization, traditional non-generative machine learning-based design, generative-artificial intelligence (AI) approaches, and metaheuristic algorithms, which reduce manual intervention, accelerate the design process, and enhance the quality of the resulting structures.

Topology optimization aims to iteratively seek for the best material layout in a given design domain subject to the loads and boundary conditions until the concerned performance is optimized [35]. Over the past thirty years, a wide range of topology optimization methods have been applied to metamaterial design [36–46]. These approaches include density-based [47], level-set [48,49], phase-field [50,51], etc. However, Topology optimization methods are inherently one-at-a-time approaches that primarily focus on designing materials with specific target mechanical properties. For example, they often aim to achieve objectives like a negative Poisson's ratio or the maximization of bulk or shear modulus. This single objective approach limits their ability to explore different microstructures to achieve a wide range of material properties [33,48]. Moreover, topology optimization methods are likely to stuck in local optimal solutions, making microstructural design highly sensitive to the initial guess.

Traditional non-generative machine learning approaches have made significant strides in advancing metamaterial design by optimizing material properties and structures. For instance, Bayesian machine learning has been applied to design supercompressible metamaterials by efficiently exploring design spaces and predicting material behaviors, all while managing uncertainties in a data-driven manner [52]. Machine learning-based surrogate models have also been utilized for inverse design, as demonstrated in the case of spinodal metamaterials [53], where they establish a direct relationship between topological parameters and material properties. Furthermore, deep learning techniques, particularly convolutional neural networks (CNNs), have been embedded within topology optimization algorithms to predict the density values for each element in the design space. [54]. Moreover, CNNs have also been used to forecast deformation behaviors and optimize unit cell geometries for materials with negative Poisson's ratios, which are increasingly applied in fields like soft robotics and actuators [55]. While these approaches have expanded the scope of design possibilities, they remain constrained by predefined design spaces and underlying assumptions, underscoring the necessity for more adaptable and flexible methodologies that can transcend these limitations.

With the rapid advancement of machine learning technology, many researchers shift their focus to generative AI methodologies for metamaterial design [56–67]. For example, Generative models like Variational Autoencoders (VAEs) can use a two-dimensional metamaterial database, with the encoder compressing the structures into a latent space vector [29,58]. By adding random Gaussian noise to the latent space and then decoding it with the decoder, new metamaterial structures are reconstructed. This process, executed in seconds with GPU time, is thousands of times faster than traditional topology optimization. Besides, Recent studies have leveraged diffusion-based generative AI models to address the inverse design problem in metamaterials. Video denoising diffusion models were trained on full-field mechanical data to predict nonlinear stress-strain responses and deformation paths [68], surpassing previous methods limited to linear properties and simple geometries. In parallel, this approach was extended by integrating multi-material structures [69], using a diffusion model and structure identifier networks to generate designs based on target nonlinear stress-strain curves. One of the most important advantages of these methods is their ability to learn representations of data with multiple levels of abstraction [70]. However, the effectiveness of a deep learning algorithm is highly dependent on the integrity of the input-data representation [71,72]. In complex tasks that lack comprehensive databases, models risk becoming biased, which limits the exploration of the design space [73,74]. Furthermore, these approaches are primarily limited to exploring a design space confined within the dataset for which they have been trained. Consequently, the creation of novel designs remains a difficult task in these efforts, owing to the difficulty to extrapolate material microstructures [75].

On the other hand, metaheuristic methods such as genetic algorithm (GA) is known for its ability to effectively explore design spaces, and has been successfully applied to minimum compliance optimization [76,77], heat transfer optimization [78], and piezoelectric structure topology optimization [79]. One of the primary advantages of GAs is their ability to handle complex multi-objective problems where gradient-based methods fail due to difficulties in calculating or deriving gradients [80]. They can overcome the common issue of traditional TO methods getting stuck in local optima, and can get a series of structures by identifying the Pareto fronts for multiple objectives [81]. Another significant advantage of GAs is that they do not require explicit definitions of material-free regions, thus avoiding grid-related issues, such as the checkerboard problem of the SIMP method [27],[83]. However, these algorithms do not scale efficiently with increasing dimensionality of the design space. The encoding scheme of metamaterial structures within GAs presents a substantial challenge. The inherent diversity and complexity of metamaterial structures complicate the direct encoding of designs, such as traditional pixelated encoding [84]. Additionally, ensuring structural integrity is also challenging, as each bit in the encoding operates independently without correlation, making it difficult to create cohesive and well-connected structures [85]. Therefore, indirect encoding methods, such as geometry-parameterized encoding, are often used to reduce the encoding dimensionality by abstracting the representation of complex structures into manageable parameters. However, these methods significantly confine the design space and require a carefully developed parameterization strategy [86].

The aim of this study is to provide an automatic, gradient-free, data-free and highly scalable framework for metamaterial design and metamaterial database construction, tackling the aforementioned challenges simultaneously. Specifically, we use the Compositional Pattern Producing Networks (CPPNs) [87], a special type of artificial neural network for pattern generation, to encode the structures of our metamaterials. We highlight that CPPNs enable scalable mapping of arbitrarily high-dimensional design spaces using very few design variables [88], which correspond to the network parameters, addressing the encoding challenges of complex structures in metaheuristic algorithms. Additionally, we employ gradient-free neuroevolution algorithms to automatically evolve CPPNs without any prior data or gradient information, overcoming the need for high-quality training data and the difficulty of training data-driven models [89],[90]. We show that by integrating CPPNs with a tailored neuroevolution algorithm, it is possible to evolve a series of fully connected and precisely defined metamaterials that approximate to optimal trade-offs in mechanical properties, all without the need for postprocessing – a challenge that has hindered prior approaches [38,91,92]. Using our proposed method, we discover the trade-off between extreme elastic properties by iteratively exploring the Pareto front of different combinations of target properties under various symmetries in two-dimensions [93]. Finally, we compile data from each iteration to establish a comprehensive database of metamaterials featuring various symmetry types, suggesting that no computational resources are wasted.

The remainder of this paper is organized as follows. Section 2 outlines the proposed framework, detailing the use of CPPNs for geometry encoding and the tailored multi-objective neuroevolution algorithm for optimization. Section 3 describes the simulation setup, including elastic tensor calculations, symmetry classifications, and design family identification. We also present the results on property space exploration and database construction. Finally, in Section 4, we offer several concluding remarks.

## 2. Proposed Framework

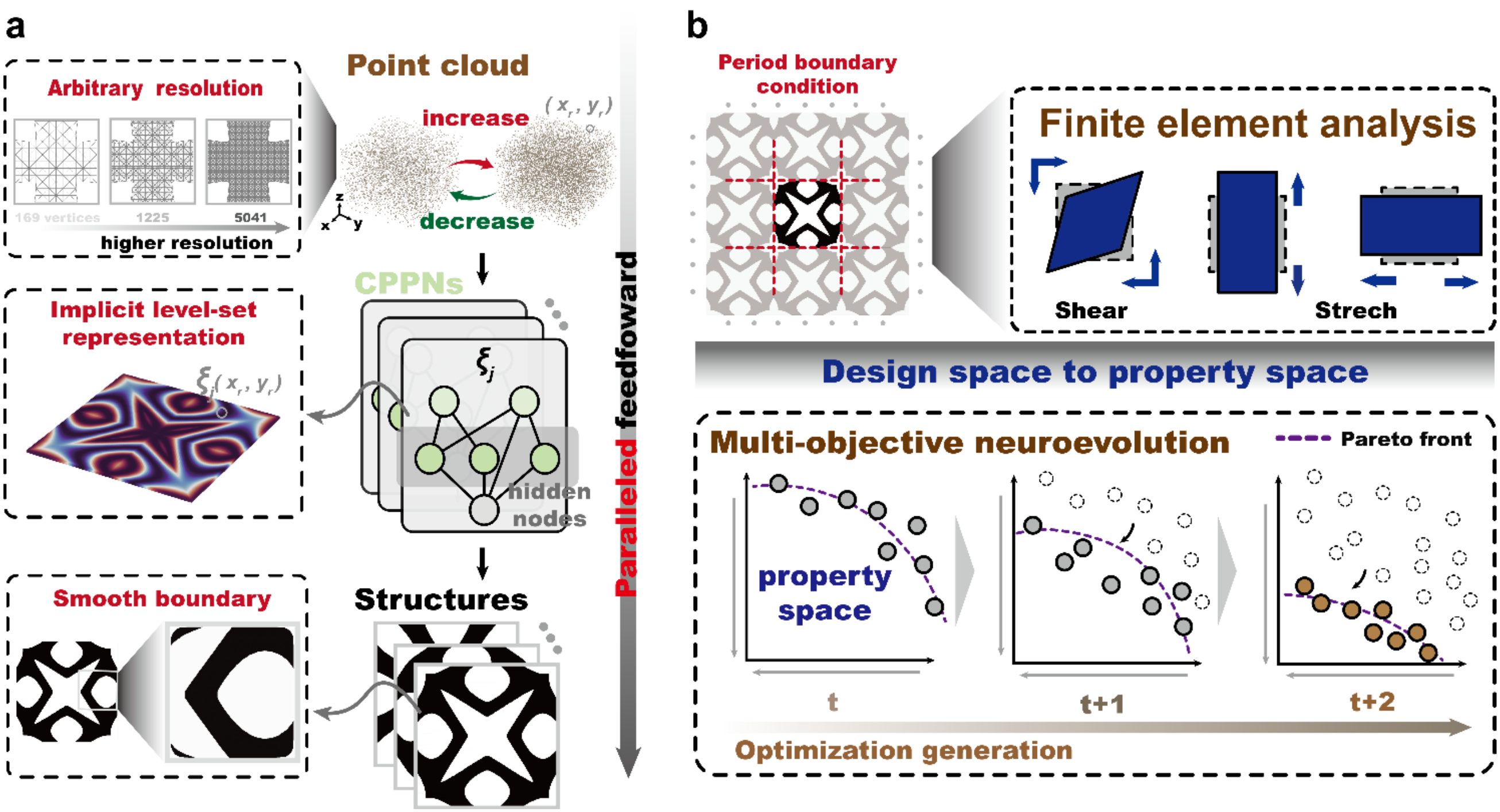

**Fig. 1** Overview of the proposed metamaterial design framework using neuroevolution. (a) The framework employs the Compositional Pattern Producing Networks (CPPNs) to encode the geometries of microstructural unit cell designs. The CPPN takes in a spatial coordinate $(x_r, y_r)$ to produce an intensity value $\xi$, which is then evaluated on a prescribed point clouds to approximate the continuous field of $\xi$. (b) Homogenized properties of the designs are evaluated via FEA with periodic boundary conditions. The optimization procedure proceeds over multiple generations, employing a customized multi-objective neuroevolution approach to find the Pareto front in property space.

We develop a CPPN-based design framework utilizing multi-objective neuroevolution to generate diverse metamaterial structures, formulating the exploration of properties trade-off as the identification of an optimal Pareto front in multi-objective optimization. This framework operates in two alternating phases: the first focuses on generating metamaterials using CPPNs, and the second employs neuroevolution to iteratively evolve the CPPNs. Both phases can be parallelized − forward propagation in CPPNs can process inputs independently, while neuroevolution allows for parallel evaluation of individuals. Such parallelization significantly accelerates the optimization process.

As illustrated in Fig. 1(a), the first phase begins with a prescribed point cloud representing the spatial distribution of material points. This point cloud serves as the foundation for extending the framework to three-dimensional structural design tasks. To generate metamaterial structures, CPPNs are employed to implicitly map the spatial coordinates of the point cloud to material distributions, encoding spatial patterns and material properties as continuous fields. For each spatial point $(x_r, y_r)$, denoted as the input $\boldsymbol{X}_r$ to the CPPN, the network evaluates an indicator function $\xi_j$, which determines whether a material occupies that point and, if so, the type of material assigned. This relationship is expressed as:

$$\boldsymbol{X}_r = (x_r, y_r), \tag{1}$$

$$\xi_j^r = \mathrm{CPPN}_j\big(\boldsymbol{X}_r; \boldsymbol{\theta}_j\big). \tag{2}$$

where $\boldsymbol{\theta}$ represents the CPPN's defining parameters.

Since each CPPN has a variable topology, they are parameterized by a graph of nodes and connections (or edges). Each node $n_k$ contains an activation function $\phi_k$ and a bias $b_k$, forming the set $n_k = \{\phi_k, b_k\}$ and collectively denoted as $N_j = \{n_1, n_2, \ldots\}$. Connections are characterized by their weights $W_{kl}$ and connectivity $c_{kl} \in \{0,1\}$, which form the set $C_j = \{(k, l, W_{kl}) \mid c_{kl} = 1\}$. The complete CPPN parameterization is thus expressed as $\boldsymbol{\theta}_j = (N_j, C_j)$. Inputs propagate through these nodes and connections to produce an intensity value $\xi_j$.

Because $\boldsymbol{X}_r$ can represent any spatial point, the $j$-th CPPN effectively encodes a continuous field of intensity value $\xi_j$ over the design domain. Structural designs are generated by applying thresholds to this field, segmenting regions into materials and voids, analogous to the level-set methods. The boundaries of the resulting designs are defined as contours of $\xi_j$. Therefore, we regard the CPPN parameters as genotype, and the resultant design evaluated from intensity values as phenotype. Typically, the field $\xi_j$ is sampled using a prescribed point cloud, which offers several key advantages. First, the resolution of the structure can be dynamically adjusted by varying the density of the point cloud, enabling scalability to high-resolution design spaces without introducing additional design variables. Second, linear interpolation can be applied to the sampled field, ensuring that the resulting material boundaries are continuous, smooth, and well-defined. This encoding approach efficiently combines flexibility in geometric representation with computational scalability, making it particularly suitable for high-resolution structural design tasks.

In the second phase, a multi-objective neuroevolution algorithm is employed to optimize the CPPN parameters $\boldsymbol{\theta}$, and thus improving the structural design of metamaterials. The workflow of the evolutionary process is illustrated in Fig. 1(b). The optimization begins by evaluating the first generation of randomly generated metamaterial structures using FEA. Structural properties such as Young's modulus $E$ and Poisson's ratio $\nu$ are calculated for each candidate design and used as fitness metrics in the MOO process. For example, given a design candidate $D_j$ with corresponding structural properties $E_j$ and $\nu_j$, the fitness of the candidate is defined as:

$$\boldsymbol{f}(D_j) = (E_j, \nu_j). \tag{3}$$

Within the property space, the optimization algorithm seeks to maximize or minimize these fitness metrics simultaneously by identifying solutions on the Pareto front $PF$ [80,94]. This process involves employing a reference-vector-guided selection strategy during each iteration [95], which uses reference vectors to guide the preservation or removal of CPPN individuals based on their proximity to different regions of the Pareto front. The Pareto front is thus formulated as:

$$PF = \{\boldsymbol{f}(D_j) \mid D_j \text{ is retained by the reference-vector-guided selection strategy}\}. \tag{4}$$

The neuroevolutionary approach operates by iteratively updating the CPPN parameters $\{\boldsymbol{\theta}_j\}$ through genetic operators, such as mutation and crossover, to explore the design space. This approach does not require fitness gradient information and is formulated as:

$$\boldsymbol{\theta}_j^{t+1} = \mathcal{G}(\boldsymbol{\theta}_j^t, \mathcal{M}, \mathcal{C}), \tag{5}$$

where $\mathcal{G}$ represents the evolutionary algorithm, $\mathcal{M}$ denotes the mutation operator, and $\mathcal{C}$ denotes the crossover operator. By avoiding reliance on gradients, this method ensures robustness against challenges such as noise, non-convexity, and discontinuities in the fitness function, which are prevalent in metamaterial design [39].

The reference-vector-guided selection plays a crucial role in managing the balance between convergence and diversity during the optimization. Reference vectors $V_0 = \{v_{0,1}, v_{0,2}, \dots, v_{0,N_v}\}$ partition the property space into regions, enabling the selection process to prioritize solutions that enhance diversity along the Pareto front. At each iteration, the algorithm adapts these vectors to improve their alignment with the Pareto front, as described in Algorithm 1. The resulting Pareto front provides a set of trade-off solutions balancing multiple mechanical properties, such as stiffness and auxeticity. The two-phase design process is summarized as follows:

**Algorithm 1** Main process of the neuroevolution framework

| | |
|---|---|
| 1: | **Input:** the maximal number of generations $t_{max}$, a set of $N_v$ initial unit reference vectors $V_0 = \{v_{0,1}, v_{0,2}, \dots, v_{0,N_v}\}$; |
| 2: | **Output:** Final optimized population of CPPNs with parameters $\boldsymbol{\theta}$; |
| 3: | /* Initialization */ |
| 4: | **Initialization**: create the initial population $P_0 = \left\{\boldsymbol{\theta}_1^0, \boldsymbol{\theta}_2^0, \dots, \boldsymbol{\theta}_{N_\mathrm{p}}^0\right\}$, where each $\boldsymbol{\theta}_1^0$ represents the parameters of a randomized CPPN individual. |
| 5: | /* Main Loop */ |
| 6: | **while** $t < t_{max}$ **do** |
| 7: | $\quad\boldsymbol{\theta}_{\mathrm{offspring}} = \mathcal{M}\left(\mathcal{C}(\boldsymbol{\theta}_{j_1}, \boldsymbol{\theta}_{j_2})\right), \boldsymbol{\theta}_{j_1}, \boldsymbol{\theta}_{j_2} \in P_t$; |
| 8: | $\quad Q_t = \{\boldsymbol{\theta}_{\mathrm{offspring}}\}$; |
| 9: | $\quad P_t = P_t \cup Q_t$; |
| 10: | For each CPPN $\boldsymbol{\theta}_j \in P_t$, generate the corresponding structural design and evaluate its performance using FEA to obtain $\boldsymbol{f}(D_j)$. |

$P_{t+1}$ = reference-vector-guided-selection$(t, P_t, V_t, \boldsymbol{f}(D))$;

$V_{t+1}$= reference-vector-adaptation$(t, P_{t+1}, V_t, V_0)$;

$t = t + 1$;

**end while**

---

It is important to note that the process of metamaterial geometry generation is decoupled from the optimization process, and the optimization process itself is decoupled from specific tasks. Firstly, due to the capability of CPPNs to indirectly encode a wide variety of geometric structures, different design requirements can be met by guiding the CPPNs to evolve in a specific direction, thereby obtaining the desired design. Secondly, the optimization process directly optimizes the CPPNs, which indirectly optimizes the resulting structure. This decoupling ensures that the optimization process is independent of the specific task. Consequently, when confronted with different design tasks, our framework could show significant flexibility and robustness. By modifying the simulation algorithm for each task and providing the corresponding performance evaluation metrics, the framework can be automatically applied to generate desired structural geometries.

The next sections provide a detailed explanation of the CPPNs and the evolution mechanism of the CPPNs through our customized neuroevolution framework.

## 2.1 Geometry encoding with CPPNs

In this section, we illustrate how CPPNs are used to generate metamaterial microstructural designs.

### 2.1.1 Capabilities of CPPNs

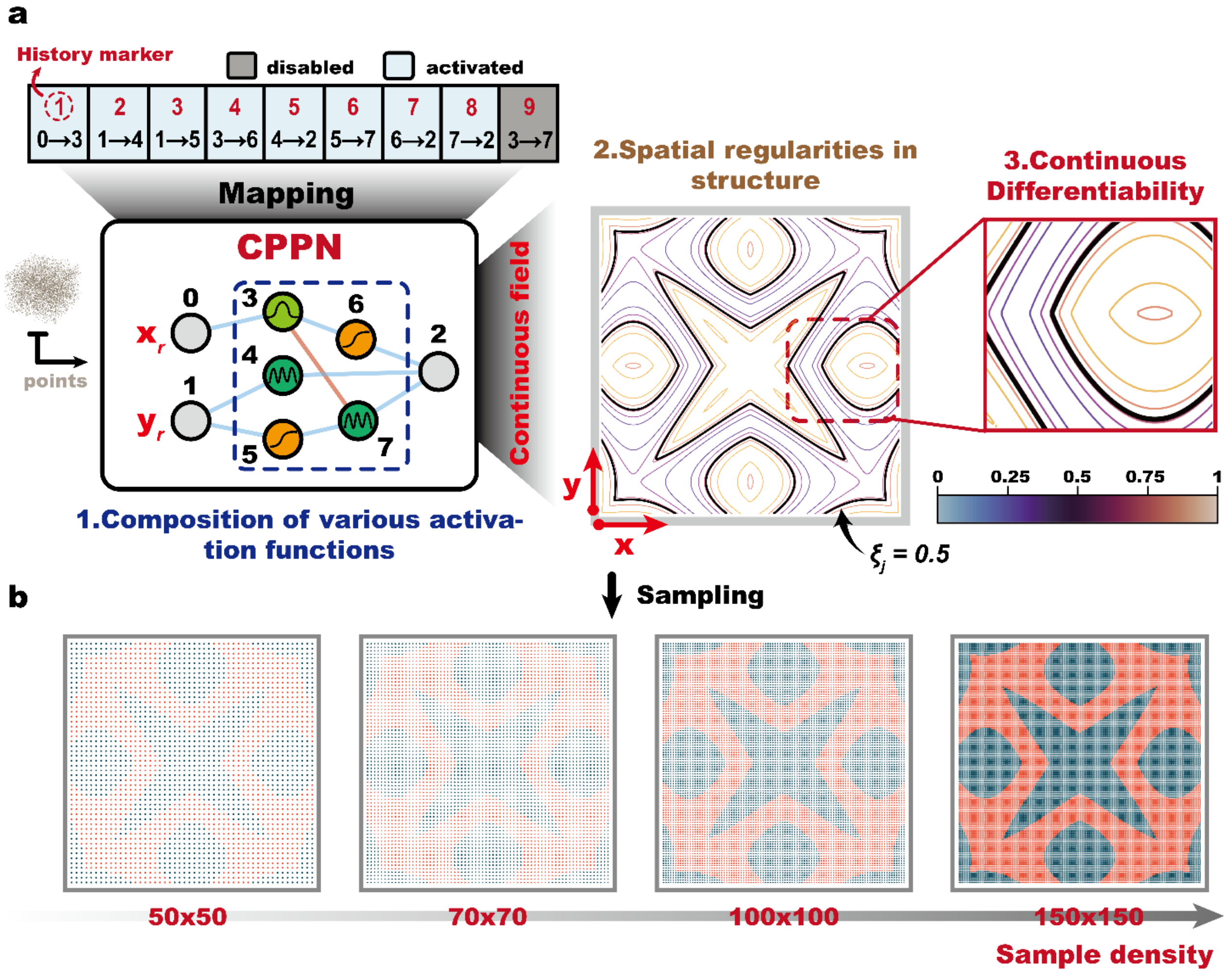

**Fig. 2** Illustration of the CPPN and its advantages in encoding geometries. (a) The network topology of a CPPN is encoded into a set, where each connection corresponds to a unique history marker, facilitating crossover and mutation operations in the subsequent evolutionary process. A CPPN can be viewed as a continuous field in space, which represents the geometry in a lossless manner. The output values of the CPPN are normalized into [0,1], allowing for the subsequent partitioning of the field using contour lines with a threshold of 0.5. (b) Once the network topology of the CPPN is determined, the continuous field it represents is sampled. By increasing the sampling density, we can generate metamaterial geometries with progressively higher resolution, with the orange regions representing the material areas and the blue regions corresponding to the void areas.

CPPNs are a specialized class of artificial neural networks designed to generate high-resolution and complex patterns through the composition of various activation functions. Given an arbitrary point in space $\boldsymbol{X}_r$, a CPPN outputs a corresponding intensity value $\xi(\boldsymbol{X}_r)$. In this manner, a structure can be implicitly encoded by a level set of the continuous field of $\xi$ [96]. Unlike traditional artificial neural networks with a fixed topology that are optimized using gradient descent, CPPNs are evolved through a neuroevolution approach that adjusts both the CPPN's weights and its topology.

As illustrated in Fig. 2, CPPNs offer several distinct advantages for structural design generation. First, CPPNs incorporate a variety of activation functions within their nodes. By leveraging this diversity of activation functions, CPPNs are able to represent highly complex geometries with only a small number of parameters, making them both efficient and powerful. Second, CPPNs tends to generate geometries with inherent spatial regularities [87]. Since CPPNs encode geometries implicitly, modifications to the CPPN lead to global changes in the resulting structures, rather than small, localized variations. Additionally, the use of continuous and differentiable activation functions enhances the smoothness and continuity of the generated geometries. For example, using a sine function in the input layer produces outputs with periodic variations, while employing a square function can generate spatially symmetric patterns. This ability enables the designs generated by CPPNs to exhibit better manufacturability.

Formally, the CPPN output can be written as:

$$h_k = \phi_k\left(\sum_{l\in\mathrm{Pre}(k)} W_{lk}\cdot h_l + b_k\right), \tag{6}$$

$$\xi_j(\boldsymbol{X}_r) = h_{out}, \tag{7}$$

where $\phi_k$ denotes a certain choice of activation function assigned to network nodes, $W_{lk}$ are the weights assigned to network connections, and $b_k$ denote nodal biases. Together, along with the network topology, they form the CPPN defining parameters $\boldsymbol{\theta}_j$. The set $\mathrm{Pre}(k)$ represents the set of all predecessor nodes pointing to node $k$, $h_k$ represents the output of node $k$ within the network, and $h_{out}$ is the value of the final output node. The choice of differentiable activation functions $\phi$ (e.g., sin, Gaussian) ensures that $\nabla\xi(\boldsymbol{X}_r)$ is continuous, so that the level set that separates different regions is free of abrupt changes, as discontinuities in the gradient are avoided. This characteristic is particularly important for generating geometries with smooth and well-defined boundaries, which is essential for mechanical metamaterials to avoid stress concentration.

Most importantly, since a CPPN represents a continuous field in space, normalizing its output to the range [0,1] allows us to partition the field and obtain precise implicit geometric representations. In this way, CPPNs can represent our metamaterial structures in a lossless manner. By increasing the number of sampling points in this field, CPPNs enable the generation of geometries at arbitrary resolution. This scalability allows for the creation of high-resolution patterns without the need to increase the size of the network. The total computational cost is proportional to the number of sampling points. Furthermore, by storing only the network parameters $\boldsymbol{\theta}_j$ , structures at any resolution can be reconstructed efficiently without significant memory overhead, as the storage size is proportional to the cardinality of the network parameters. This feature makes CPPNs highly memory-efficient for large-scale databases. For example, in this research, after applying lossless compression, a database containing 486852 samples only cost 5.7GB of memory, with each individual consuming approximately 12KB.

### 2.1.2 Encoding scheme of CPPNs

Internally, a CPPN is represented as a directed acyclic graph (DAG) of functions. The output value of a function node is multiplied by the weight connecting it to the input of the next function node. If multiple input connections feed into the same function node, then that node takes the sum of its weighted input values as its input. As shown in Fig. 2, each CPPN contains a set of connection genes and a set of node genes. Each connection gene specifies several key parameters: the input and output nodes, the connection weight, an enable bit indicating whether the connection is active, and a unique historical marker that tracks the gene's lineage. Each node gene defines the node's internal activation function, selected from a predefined list of functions such as linear, sigmoid, Gaussian, sine, and cosine. Specially, to avoiding the confliction among different CPPNs during the evolution process, we implement a global historical marking system that directly tags genes within the CPPNs. Whenever a new gene is created, it is assigned a marker that corresponds to its chronological order of appearance in the evolving population. When genes are reproduced and transmitted to offspring, they retain their original markers. This system ensures consistency in node and connection tracking across generations, preventing duplication and preserving the integrity of the neuroevolution process. In doing so, it also lays the groundwork for subsequent similarity assessments of CPPNs, making the identification of metamaterial families possible.

### 2.1.3 Metamaterial topology generation

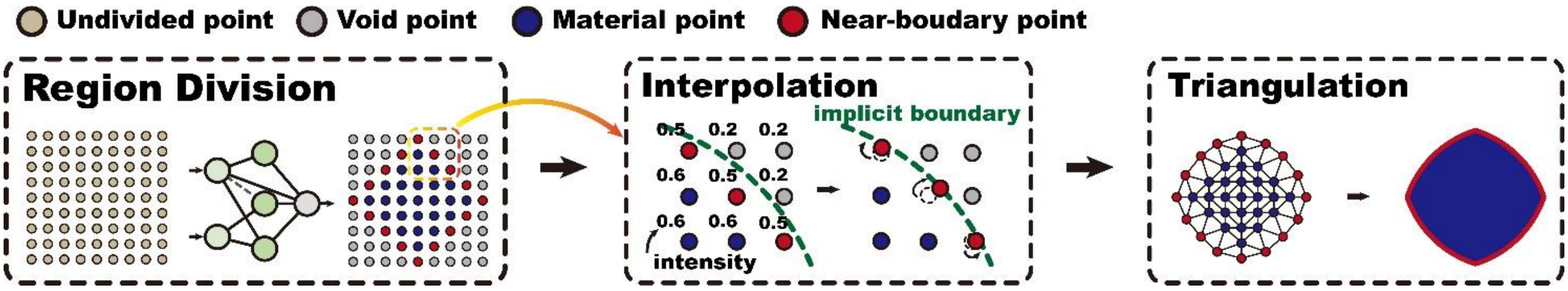

**Fig. 3** Discretization of CPPN output to FE mesh through point cloud. First, a CPPN assigns scalar values to the prescribed point cloud, classifying them into void, material, and near-boundary points. Next, a linear interpolation scheme is applied in order to adjust the locations of the near-boundary points so that they are snapped to the boundary. The adjusted point cloud is then converted to triangular FE mesh through Delaunay triangulation.

The process of generating metamaterial structures is illustrated in Fig. 3. In our task, we decode the metamaterial microstructure using point cloud, where the spatial coordinates of each point serve as inputs to the CPPN. After processing the data as described, we obtain an intensity value for each point in the domain. These intensity values are then normalized using Min-Max scaling to constrain the range between 0 and 1.

$$\overline{\xi_j^r} = \frac{\xi_j^r - \xi_{\min}}{\xi_{\max} - \xi_{min}}, \tag{8}$$

Where $\xi_{\max}$ and $\xi_{\min}$ represents the minimum and maximum intensity values produced by the CPPN defined by $\boldsymbol{\theta}_j$ for the given point clouds, respectively. The variable $\xi_j^r$ denotes the original intensity value of the $r$-th point in point clouds. This normalization ensures that all intensity values are linearly scaled within the interval $[0,1]$, preserving their relative magnitudes while providing a consistent range for further computations and visualizations.

The next step involves both classifying the material and non-material regions based on the normalized intensity values and identifying the contour that separates these regions. Specifically, a threshold $T$ of 0.5 is applied to the normalized intensity values to divide the point cloud into two regions: material regions and non-material regions. To detect the boundary between these regions, we implemented a straightforward boundary detection method. In this method, the two regions are labeled as -1 and 1, respectively, to facilitate the identification of contour points:

$$Label_j(x_r, y_r) = \begin{cases} 1, & \xi_j^r \geq T, \\ -1, & \xi_j^r < T, \end{cases} \tag{9}$$

where $\xi_j^r$ is the intensity value at $r$-th point evaluated by the $j$-th CPPN, and $T$ is the threshold (e.g., $T = 0.5$).

To identify boundary points, we first identify existing points that are close to the implicit boundary. Then we apply linear interpolation to each pair of points that are on the opposite sides of the boundary to obtain explicit boundary points [97]:

$$\boldsymbol{X}_b = \rho \cdot \boldsymbol{X}_p + (1-\rho) \cdot \boldsymbol{X}_q, where\ \rho = \frac{\left|\xi_j^p - T\right|}{\left|\xi_j^p - \xi_j^q\right|}. \tag{10}$$

Where $\boldsymbol{X}_p$ and $\boldsymbol{X}_q$ are neighboring points with $\xi_j^p > T\ and\ \xi_j^q < T$, $\boldsymbol{X}_b$ denotes the boundary point between points $\boldsymbol{X}_p$ and $\boldsymbol{X}_q$.

Finally, we need to convert the point cloud data into a triangular mesh to facilitate subsequent finite element analyses. This conversion is done by applying the Delaunay triangulation algorithm to the previously obtain point cloud that are label by 1, including the boundary points. By doing so, we can quickly generate a valid triangular mesh [98].

## 2.2 Multi-objective neuroevolution algorithm

In this section, we discuss how to evolve CPPNs to generate desired structures, as shown in the Fig. 4. We provide an in-depth explanation of the execution process, addressing key aspects such as reproduction, selection strategies, and the challenges of managing constraints in metamaterial design generation.

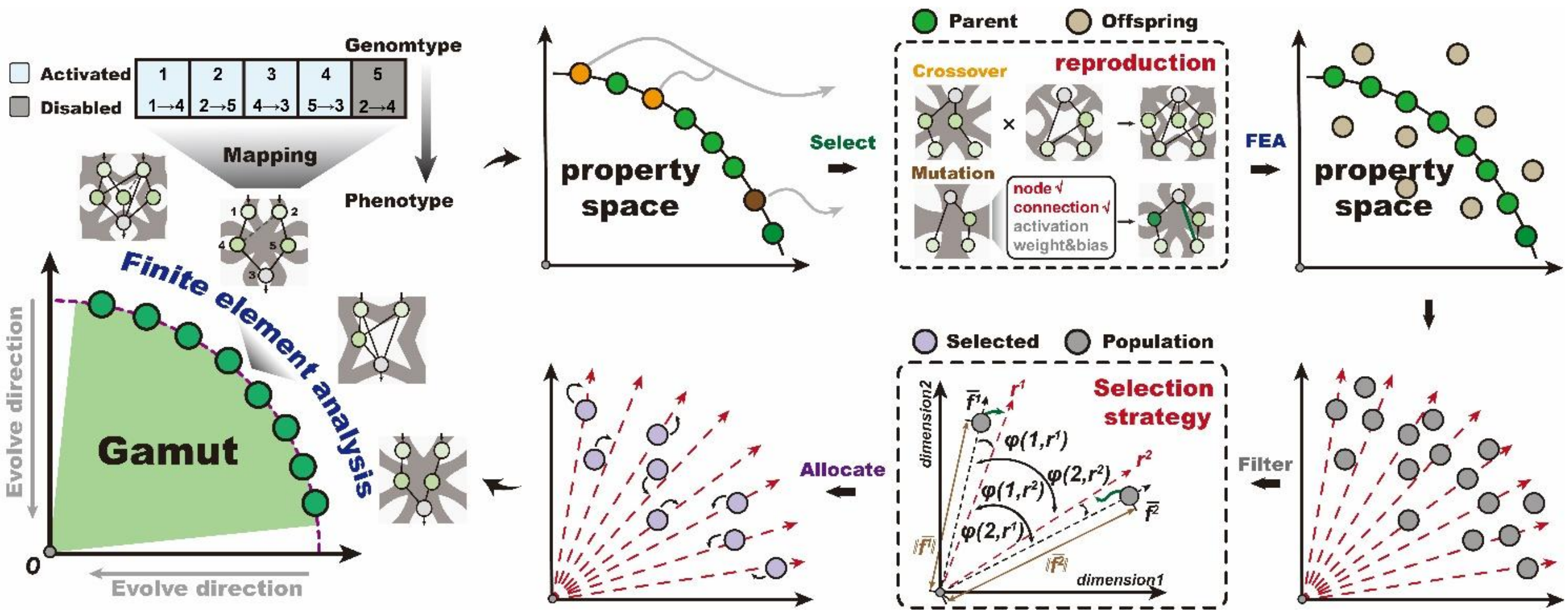

**Fig. 4** Evolution process for metamaterial design. The genotype (i.e., $\boldsymbol{\theta}$) encodes the CPPN network structure, which maps to a phenotype representing a metamaterial design. FEA evaluates the design's performance in the property space defined by adjusted fitness metrics $f'$, forming the property gamut. Reproduction via crossover and mutation generates offspring by modifying the genotype, including nodes, connections, activation functions, and weights. A reference-vector-guided selection strategy identifies the current Pareto-optimal solutions, steering the population toward improved properties.

### 2.2.1 Reproduction

After utilizing CPPNs for the implicit encoding of metamaterial microstructures, it becomes crucial to establish a method that effectively directs the evolution of these networks towards generating structures with desirable mechanical properties. To achieve this, we employ neuroevolution [99], a technique that optimizes the architecture and weights of neural networks using evolutionary algorithms. Drawing inspiration from natural evolution, neuroevolution simulates processes such as crossover, mutation, and selection to progressively enhance the performance of neural networks.

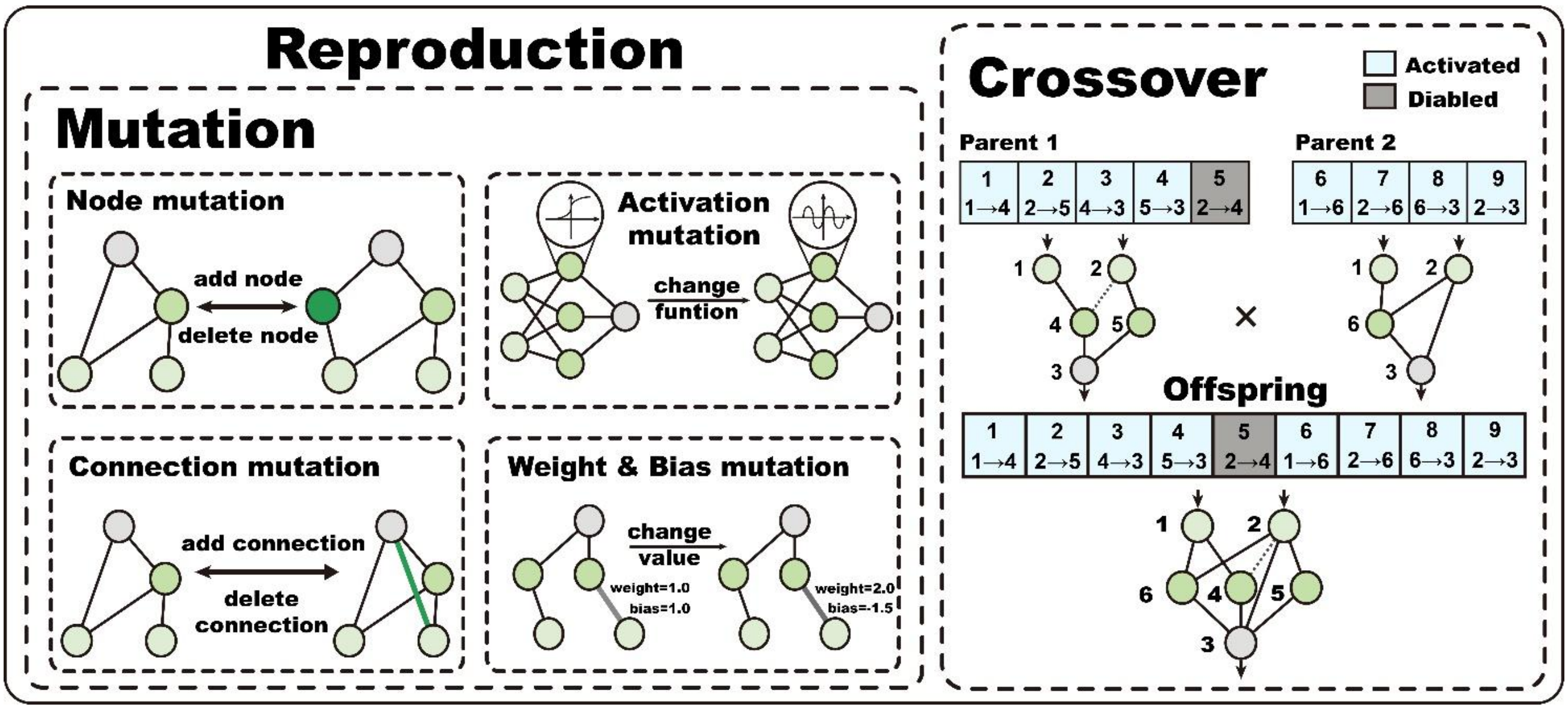

**Fig. 5** The reproduction operations for CPPNs, involving both crossover and mutation operations. The mutation phase introduces diversity through four mechanisms: node mutation (adding or removing nodes), connection mutation (rewiring connections), activation mutation (changing node activation functions), and weight & bias mutation (adjusting parameters). During crossover, each parent's genotype is represented as a dictionary, in which every connection and node is assigned a unique global historical marker. These markers serve as a universal reference system, enabling the offspring to accurately align and inherit corresponding networks from parents.

(1) For crossover, we employ concepts from the NEAT algorithm, utilizing the aforementioned marker system to align genes with the same historical origin across diverse topologies (Fig. 5). The crossover operation merges the shared traits of parents by taking the union of the genotypes of two different selected CPPNs as shown in the picture.

(2) Mutation evolves neural networks through various mechanisms, including weight mutation (perturbing or resetting connection weights), connection mutation (adding new connections between previously unconnected nodes), and node mutation (splitting existing connections to introduce new nodes, disabling the original connection, and adding two new ones).

Alterations at the genotype level often result in substantial structural changes, possibly leading to insufficient exploration of the design space, which in turn causes blanks along the target Pareto front. Therefore, we optimize the population through two phases. The first phase is the diversity exploration process, occurring in the early iterations, which integrates global information through crossover and mutation to achieve structural diversity in metamaterials. The second phase, occurring in the mid to late iterations, focuses on property space exploration through local and global information integration. In this phase, adjacent reference vectors are clustered, dividing the population into multiple niches. Local information integration restricts crossover and mutation within each niche, avoiding significant structural changes and ensuring fully occupied Pareto fronts.

---

**Algorithm 2** Reproduction in CPPNs

---

**Input:** the current generation $t$, max generation $t_{max}$, a set of parent population $P_t$.

**Output:** offspring population $Q_t$

/* Calculate Local and global reproduction numbers */

$N_{local} = \left\lfloor \frac{|P_t| \times t}{t_{max}} \right\rfloor$ // Number of individuals for local reproduction

$N_{global} = |P_t| - N_{local}$ // Number of individuals for global reproduction

$Niche$, Number$_{niche}$ = merge-niche ($P_t$,$N_{local}$);

Initialize $Q_t = \emptyset$; // Offsprings

/* Local reproduction */

**for** $i$ = 1 to | $Niche$| **do**

    **for** $j$ = 1 to Number$_{niche}$ $[i]$ **do**

        child = create-new ($Niche[i]$);

        $Q_t = Q_t \cup \{child\}$;

    **end for**

**end for**

/* Global reproduction */

**for** $k$ = 1 to $N_{global}$ **do**

    child = create-new ($P_t$);

    $Q_t = Q_t \cup \{child\}$;

19: **end for**

## 2.2.2 Selection strategy

After the crossover and mutation processes, we obtain a new generation of CPPNs, which are not necessarily superior to the parent generation. Following the finite element analysis of each individual, selection among parents and offspring is conducted using the RVEA strategy [42].

First, a set of uniformly distributed reference vectors $V = \{v_1, v_2, \dots, v_{N_v}\}$ is generated to represent different directions in the objective space, where each $v_i$ is normalized such that $\|v_i\| = 1$. In this context, each solution $f_j$ represents the fitness value of a metamaterial mapped by the $j$-th CPPN, with the fitness evaluated by FEA. RVEA then projects each solution $f_j$ onto these reference vectors and assigns solutions to the vectors based on these projections. The projection is determined by minimizing the angle $\varphi_{ij}$ between the normalized objective vector $f'_{t,j}$ and the reference vector $v_{t,i}$:

$$\varphi_{t,i,j} = \arccos\left(\frac{f'_{t,j} \cdot v_{t,i}}{\| f'_{t,j} \|}\right). \tag{11}$$

The normalized objective values $f'_{t,j}$ are computed using the ideal point $z_t^{\min}$ :

$$z_t^{\min} = \min_{j \in \{1,2,\dots,|P_t|\}} f_{t,j} , \tag{12}$$

$$f'_{t,j} = f_{t,j} - z_t^{min}. \tag{13}$$

During selection, RVEA prioritizes solutions that perform better within their respective reference vector groups by using the Angle Penalty Distance (APD) values, where a smaller APD value indicates a superior solution. The APD for a solution $f_{t,j}$ relative to a reference vector $v_{t,i}$ is given by:

$$d_{t,i,j} = APD(f_{t,j}, v_{t,i}) = \| f'_{t,j} \| \cdot \left(1 + P(\varphi_{t,i,j})\right). \tag{14}$$

$$P(\varphi_{t,i,j}) = M \cdot \left(\frac{t}{t_{max}}\right)^{\alpha} \cdot \frac{\varphi_{t,i,j}}{\gamma_{v_{t,i}}} , \tag{15}$$

where $d_{t,i,j}$ is the APD values from the solution to the reference vector, $\varphi_{t,i,j}$ is the angle between the solution and the reference vector, $t$ is the current iteration number, $t_{\max}$ is a pre-defined maximum iteration number, $\alpha$ is a user defined parameter controlling the rate of change of $P(\varphi_{t,i,j})$, $M$ is the number of objectives, and $\gamma_{v_{t,i}}$ is the smallest angle value between reference vector $v_{t,i}$ and the other reference vectors in the current generation. This prioritization ensures a uniform distribution of solutions across different objective directions, maintaining diversity while promoting convergence.

---

**Algorithm 3** Reference Vector-Guided Selection Strategy

---

**Input:** generation index $t$, population $P_t$, unit reference vector set $V_t = \{v_{t,1}, v_{t,2}, \dots, v_{t,N_v}\}$, fitness values list $f_t$;

**Output**: population $P_{t+1}$ for next generation;

/* Objective Value Translation */

$z_t^{min} = \min(f_t)$; *// Find minimal values of each objective*

**for** $j = 1$ to $|P_t|$ **do**

    $f'_{t,j} = f_{t,j} - z_t^{min}$;

**end for**

/* Population Partition */

**for** $j = 1$ to $|P_t|$ **do**

    **for** $i = 1$ to $|V_t|$ **do**

        $\cos \varphi_{t,i,j} = \frac{f'_{t,j} \cdot v_{t,i}}{\|f'_{t,j}\|}$;

    **end for**

**end for**

$I_t = \emptyset$; *// Assign individuals to the nearest vector*

**for** $i = 1$ to $|V_t|$ **do**

    $I_{t,i} = \emptyset$;

**end for**

**for** $j = 1$ to $|P_t|$ **do**

  $k = \arg \max_{j \in \{1,2,\ldots,|V_t|\}} \cos \varphi_{t,i,j}$

  $I_{t,k} = I_{t,k} \cup \{j\}$

**end for**

/* Angle-Penalized Distance (APD) Calculation */

**for** $i = 1$ to $|V_t|$ **do**

  **for** $j$ in $I_{t,i}$ **do**

    $d_{t,i,j} = \left(1 + P(\varphi_{t,i,j})\right) \times \| f'_{t,j} \|$;

  **end for**

**end for**

/* Elitism Selection */

$P_{t+1} = \emptyset$; // *Initialize the next generation population*

**for** $i = 1$ to $|V_t|$ **do**

  $k = \arg \min_{(i \in \{1,2,\ldots,|P_t|\})} d_{t,i,j}$;

  $P_{t+1} = P_{t+1} \cup P_{t,k}$;

**end for**

### 2.2.3 Constraints handling

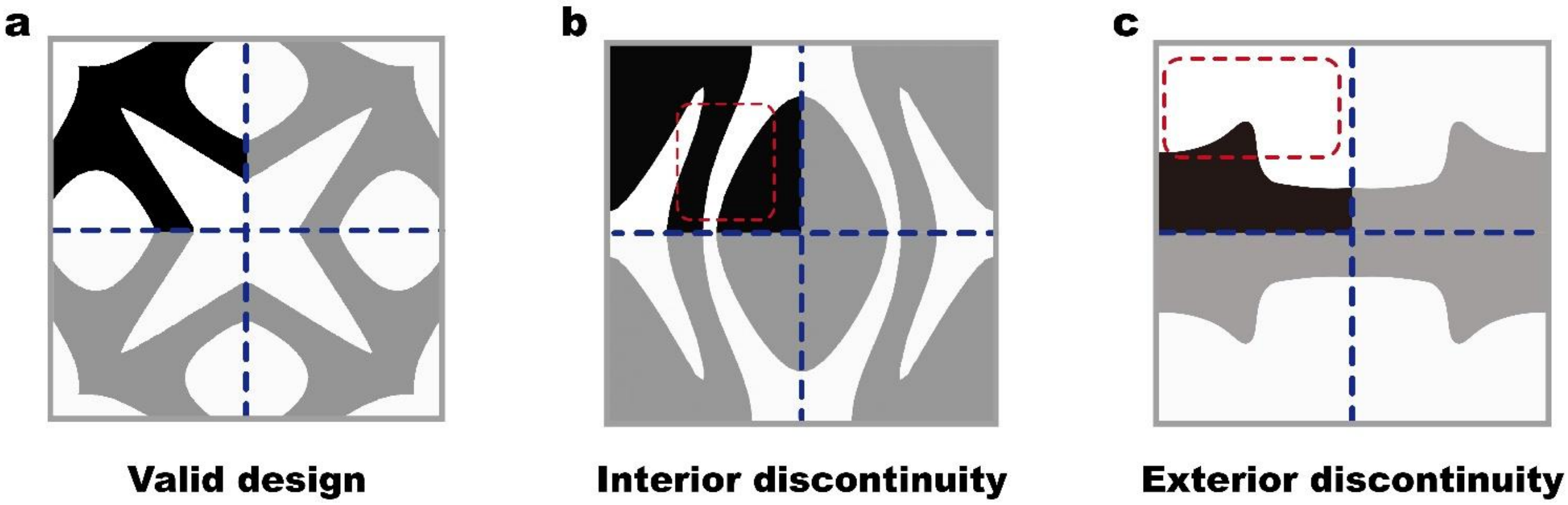

**Fig. 6** Unit cell design validation. (a) A design with no violations. (b) A design with interior discontinuity, where the unit cell itself is disconnected. (c) A design with exterior discontinuity, where the interior of the unit cell is connected, but the periodic tessellation becomes disconnected.

The structures generated by CPPNs are not without flaws, and some problematic designs cannot have their fitness calculated, as shown in the Fig. 6. The simulation of true voids can lead to issues such as discontinuous internal structures, and discontinuities when periodically tessellated. We address these problems by treating them as inequality constraints, transforming the optimization problem into a multi-objective neuroevolution under multi-constraints [100].

We stress that simply discarding structures that violate these constraints would severely limit the exploration of our objective space, thereby reducing the efficiency of design exploration [44]. Imagine an early-stage individual with a novel structure that does not yet meet the constraint conditions. While it may initially fail, this individual could evolve into a solution that satisfies the constraints after undergoing several generations of mutation. For this reason, we reject the simplistic approach of discarding individuals that fail to meet constraints from the start. Instead, we adopt an elite selection strategy based on a constraint violation function, which quantifies the number of violated constraints through a CV (Constraint Violation) metric [92].

The principle of our elite selection strategy based on the CV function is detailed as follows. First, we identify all the solutions in the current subpopulation $I_{t,i}$ that do not meet the specified constraints and record their indices in a set $NS$. We then evaluate the size of $NS$. If the size of $NS$ matches the size of the current subpopulation, indicating that every solution is infeasible, we select the solution with the minimum degree of constraint violation, calculated as the sum of the magnitudes of all constraint violations. This ensures that the least violating solution is preserved for further evolution. Conversely, if not all solutions are infeasible, we choose the one with the minimal APD value among the feasible solutions. This strategy allows us to maintain potentially promising individuals that could evolve to meet the constraints while ensuring the optimization process continues to explore novel areas of the objective space. Finally, through this constraint-handling mechanism, we decouple the treatment of design constraints from the optimization process. Specifically, when new constraints are introduced, we simply incorporate them into the CV metric calculation, and the optimization process automatically selects and eliminates individuals based on their CV values. This characteristic enhances the robustness and adaptability of our framework.

**Algorithm 4** Elitism Selection Strategy for Handling Constraints

**for** $i = 1$ to $|V_t|$ **do**
  $NS = \emptyset$;
/* Select solutions that violate the constraints */
  **for** $j$ in $I_{t,i}$ **do**
    **if** $CV(P_{t,j}) > 0$ **then**
      $NS = NS \cup \{j\}$
    **end if**
  **end for**
/* Elitism Selection */
  **if** $|NS| == |I_{t,i}|$ **then**
    $k = \arg \min_{j \in \{1,2,\ldots,|P_t|\}} CV(P_{t,j})$;
  **else**
    $k = \arg \min_{j \in \{1,2,\ldots,|P_t|\}, j \notin S} d_{t,i,j}$;
  **end if**
  $P_{t+1} = P_{t+1} \cup P_{t,k}$;
**end for**

## 2.3 Postprocessing: Family Identification via Genetic Similarity

Notably, our framework efficiently computes and identifies families of CPPNs. After completing the evolution, we evaluate the genetic similarity among all collected CPPN individuals. The genetic similarity metric is defined based on the topological similarity between two CPPNs, quantified by the number of hidden nodes, the connections between these nodes, and the weights of these connections. By clustering the CPPNs using a predefined similarity threshold, we are able to group them into distinct families.

The genetic similarity $\delta$ between different CPPNs is measured as a linear combination of three factors: the number of excess genes $G_{excess}$, the number of disjoint genes $G_{disjoint}$, and the average weight difference of matching genes $\overline{W}$, including disabled genes:

$$\delta = \frac{c_1 G_{excess}}{N_{gene}} + \frac{c_2 G_{disjoint}}{N_{gene}} + c_3 \cdot \overline{W} \tag{16}$$

Here, genes refer to the components of the CPPN that encode the network's topology, such as weights, connections and nodes. Specifically, excess genes are those that appear in one CPPN but not in another, indicating additional complexity or variation, while disjoint genes are those that appear in both networks but are not matched in corresponding positions, suggesting partial similarity or structure divergence. Matching genes are those present in both CPPNs at corresponding positions, with the average weight difference of these genes reflecting how similar the weights are between the networks. Disabled genes, which are present but inactive, are also taken into account in this calculation. The coefficients $c_1$, $c_2$, and $c_3$ adjust the importance of the three factors, and $N_{gene}$, the number of genes in the larger genome, normalizes for genome size. The similarity measure $\delta$ allows us to categorize CPPNs using a threshold $\delta_{threshold}$. Specifically, in each generation, genomes are sequentially placed into species, with each species represented by a random genome from the previous generation. A given genome $g$ in the current generation is placed in the first species where $g$ is similar with the representative genome of that species, with $\delta$ being less than the predefined $\delta_{threshold}$, ensuring that species do not overlap. If $g$ is not similar with any existing species, a new species is created with $g$ as its representative. In our task, we set $c_1 = 0.5$, $c_2 = 0.5$, $c_3 = 1$, and $\delta_{threshold} = 1.35$.

Through multiple tests, we observed that CPPNs with similar genotypes (i.e., the $\boldsymbol{\theta}$ of CPPNs) tend to produce phenotypes (i.e. metamaterial designs) with similar structural characteristics. Leveraging this feature, we efficiently organize the CPPN population into distinct families of metamaterial designs. We observe that the properties of metamaterials within the same family are widely distributed across the property space. This means that even though members of a family share structural similarities, they can exhibit a broad range of material properties. This classification simplifies the analysis and exploration of the design space, as we can focus on entire families rather than individual instances, thus facilitates the identification of underlying design principles.

## 3. Numerical Examples

In this section, we evaluate the performance of the proposed framework by examining the trade-offs between linear elastic properties of metamaterials under various symmetries. Specifically, we focus on four types of symmetries: isotropic (plane group: p31m), tetragonal (plane groups: p4mm and p4), orthotropic (plane group: p2mm), and fully anisotropic (plane group: p1) [102]. Through the evolution process, we progressively converge toward the optimal Pareto front of different combinations of material properties, allowing us to explore the limits of attainable elasticity.

For all cases, we utilized a population size of 500, with mutation probabilities of 0.2 for connection weights and 0.5 for nodal activation functions. The input point cloud is a 35×35 two-dimensional uniform grid. The evolution is performed for 800 iterations, with activation functions selected from the following options: square, sigmoid, Gaussian, sine, ReLU, and tanh.

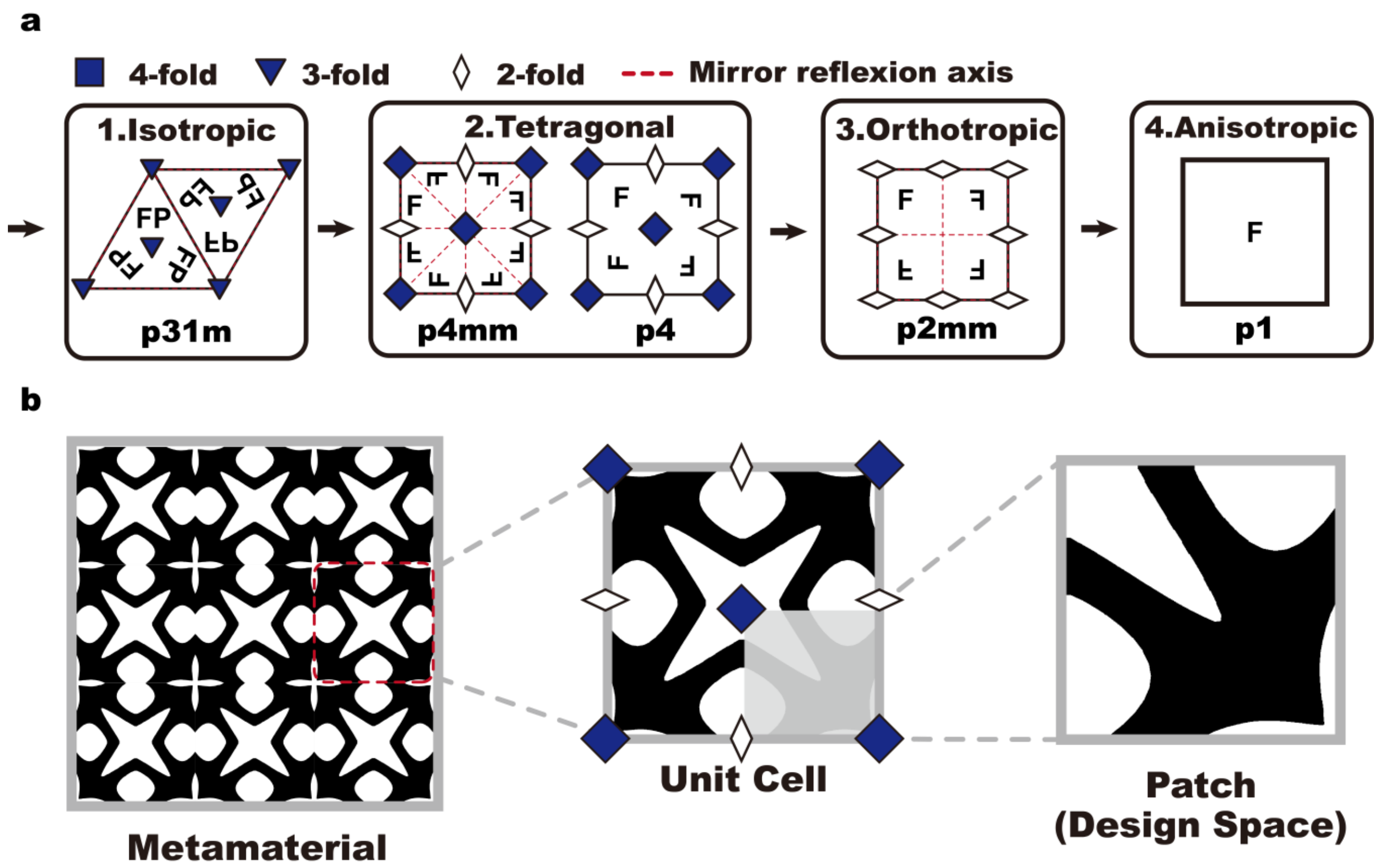

**Fig. 7** Representation of metamaterial microstructures. (a) Four types of symmetry: isotropic (p31m), tetragonal (p4mm and p4), orthotropic (p2mm), and fully anisotropic (p1). (b) The decomposition of a metamaterial design into repeating unit cells and individual patches.

Our method for generating metamaterial topologies with specified symmetry types begins by defining the topology of a basic patch (Fig. 7a). This basic patch is mapped onto a unit cell according to the predefined symmetry and then periodically tessellated to form the metamaterials. To streamline this process, we only generate the point cloud within the basic element and feed it into the CPPN to obtain their indicator values. The indicator values of other points within the unit cell are assigned according to symmetry, without CPPN evaluation.

This section is organized into five parts. First, under isotropic conditions, we explore the mechanical property space of average Young's modulus and average Poisson's ratio for designs belonging to the p31m symmetry group. Next, we examine the trade-offs in average Young's modulus and Poisson's ratio under tetragonal symmetry. Particularly, to further investigate the structural characteristics under strong symmetry (i.e., four symmetry planes), we employ a family classification based on genetic similarity. This allows us to partition the property space into distinct metamaterial families, each sharing similar structural features but exhibiting different mechanical properties. In the orthotropic case, we begin by replicating the trade-off relationships between elastic tensors as outlined in Ref. [5]. Lastly, in the anisotropic case, we pursue the shear-normal coupling effect by exploring the trade-off between elastic constants $C_{12}$ and $C_{13}$. The evaluation method of the elastic properties is detailed in Appendix A.

All computations were executed on Alibaba Cloud machines equipped with AMD EPYC 7H12 64-core processors, totaling 256 processors with a base clock speed of 2.60 GHz, and 503 GiB of DDR4 RAM. This computational setup enabled large-scale parallel processing, which is critical given the high dimensionality of the design space and the complexity of the algorithm. GPUs were not required, as the simulations primarily relied on CPU-intensive tasks to model mechanical properties.

### 3.1 Isotropic Designs

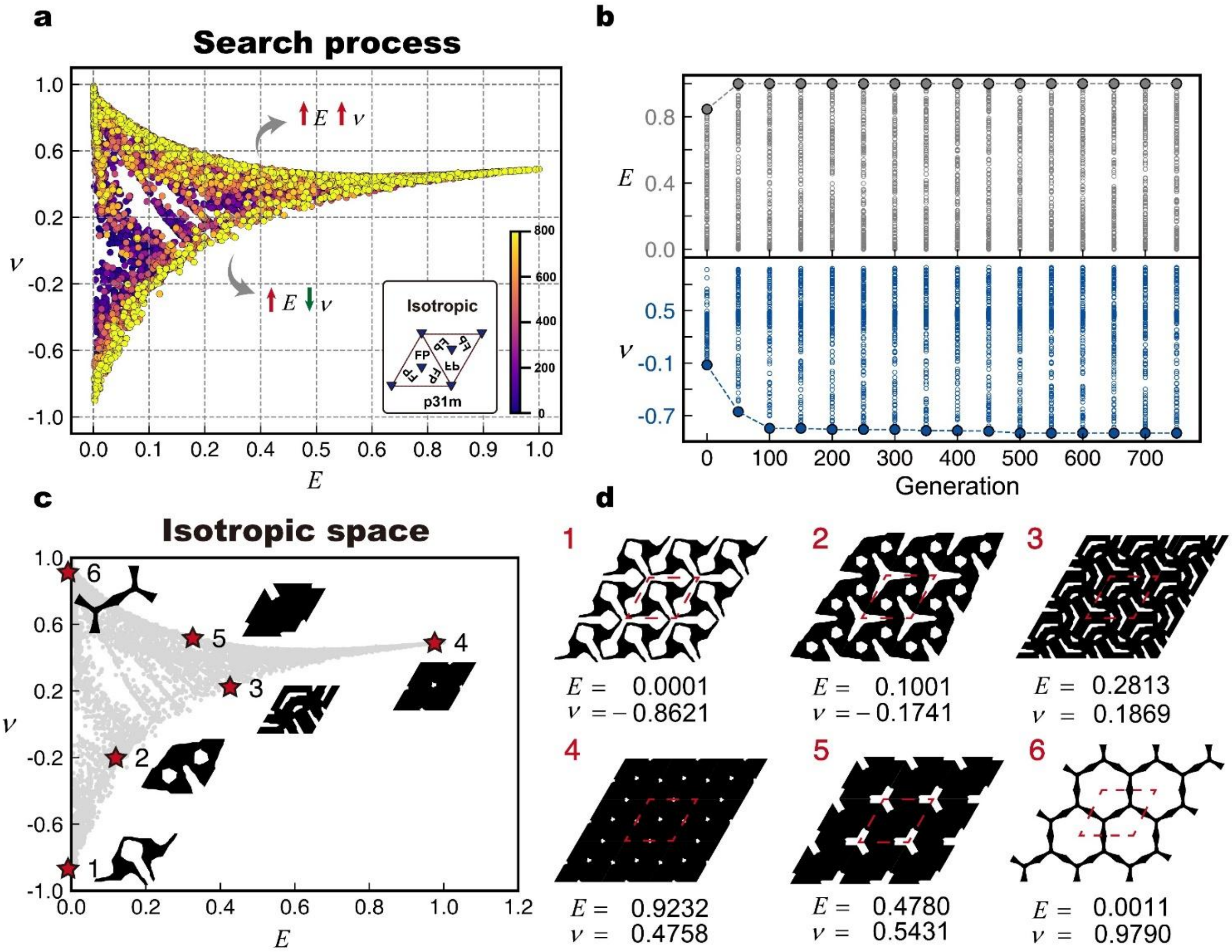

**Fig. 8** Metamaterial designs with isotropic symmetry (p31m). (a) The search process and search directions as shown in the property space of Young's modulus $E$ and Poisson's ratio $\nu$. (b) Distribution of individual properties along the evolution. (c) A representative selection of designs on the bond of properties. (d) Enlarged views of the selected designs.

Fig. 8 illustrates the exploration of the isotropic material design space using our framework. In the Fig. 8(a), the search process is visualized through the distribution of material candidates within the $E$-$\nu$ space. Under isotropic conditions in 2D, $\nu$ is theoretically bounded between -1 and 1, and our algorithm successfully approaches to the limits. Notably, our algorithm identifies not just a single solution at these extremes, but a series of designs, highlighting its ability to thoroughly search and uncover a diverse set of configurations even near the extreme bound of properties. Additionally, as observed in the Fig. 8(b), the evolution process initially concentrates on rapidly approaching the bound of the material properties. Once early convergence is achieved, the evolution process shifts its focus to pushing diversity within the population. This ensures a wide and evenly distributed set of solutions along the Pareto front, maintaining diversity across the $E$-$\nu$ space.

The analysis of the isotropic material design space reveals diverse mechanical behaviors across six distinct unit cell designs, as shown in Fig. 8(c) and Fig. 8(d), each defined by specific values of $E$ and $\nu$. Unit cell 1, with $E = 0.0001$ and $\nu = -0.8621$, demonstrates extremely low stiffness and strong auxetic behavior. Unit cell 2, characterized by $E = 0.1001$ and $\nu = -0.1741$, retains a negative Poisson's ratio while showing moderate stiffness, balancing flexibility and rigidity. Unit cell 3 transitions to conventional behavior with $E = 0.2813$ and $\nu = 0.1869$, contracting laterally under tensile stress. Unit cell 4, with $E = 0.9232$ and $\nu = 0.4758$, exhibits significant stiffness and positive Poisson's ratio, making it suitable for applications requiring both strength and flexibility. Unit cell 5 features $E = 0.4780$ and $\nu = 0.5431$, ideal for load-bearing uses. Unit cell 6, with $E = 0.0011$ and $\nu = 0.9790$, combines low stiffness with a high positive Poisson's ratio. Together, these microstructures demonstrate the evolution framework's effectiveness in exploring the isotropic design space and facilitating the development of tailored materials for various engineering applications.

## 3.2 Tetragonal Designs

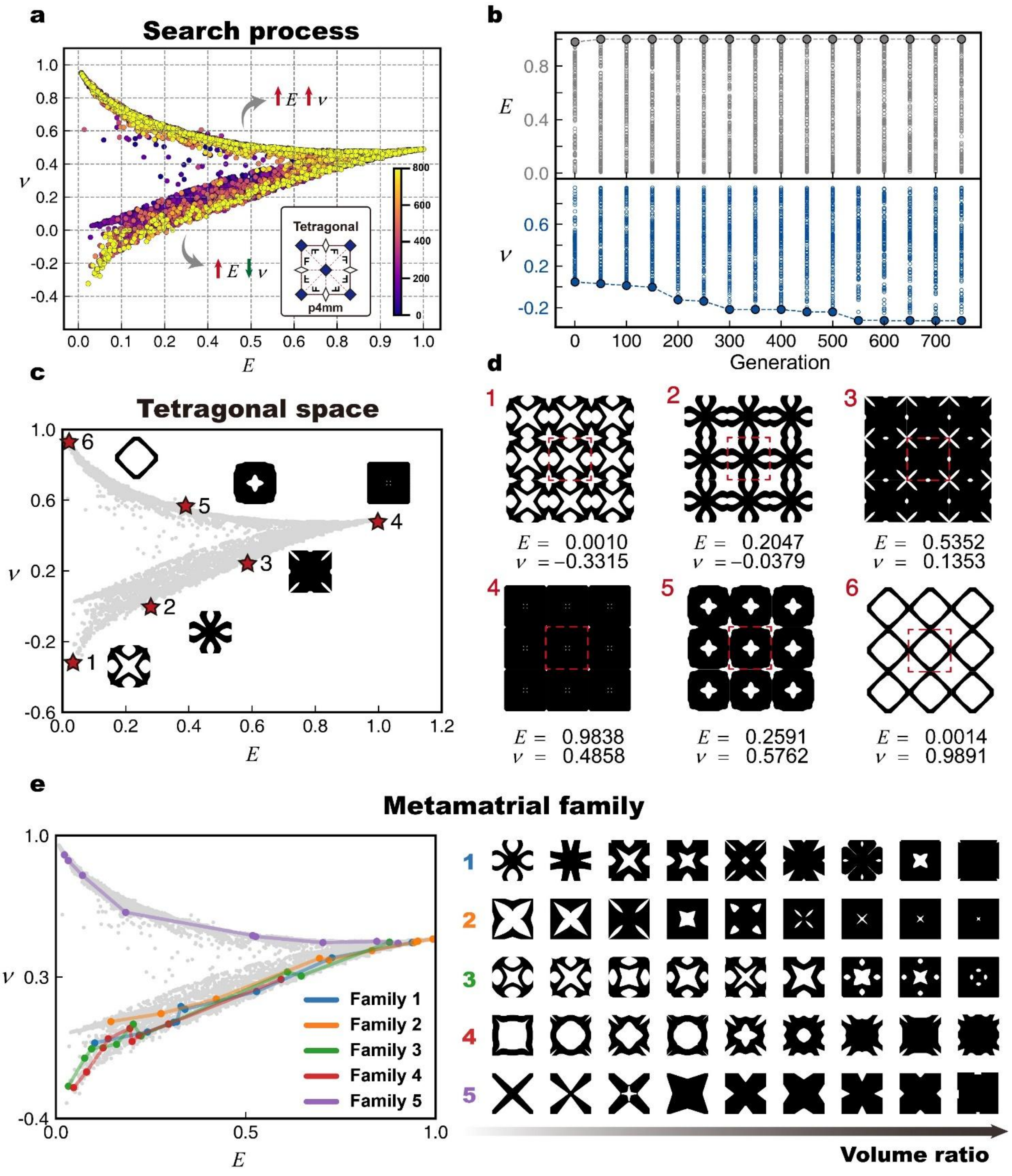

**Fig. 9** Metamaterial designs with the first kind of tetragonal symmetry (p4mm). (a) The search process and search directions as shown in the property space of Young's modulus $E$ and Poisson's ratio $\nu$. (b) Distribution of individual properties along the evolution. (c) A representative selection of designs on the bond of properties. (d) Enlarged views of the selected designs.

(e) Family identification results using the previously described similarity measure. The color indicates different family, with each class representing a group of CPPNs with similar genotypes, where the genetic similarity is measured based on the number of excess genes, disjoint genes, and the average weight difference of matching genes.

Next, we focus on exploring the tetragonal metamaterial unit cells with four symmetry planes (plane group: p4mm). Fig. 9(a)illustrates how the evolution process systematically explores the mechanical property space. The exploration proceeds in two primary directions: one aims to maximize both $E$ and $\nu$, while the other seeks to maximize $E$ while minimizing $\nu$. As shown in Fig. 9(b), the evolution of mechanical properties rapidly converges to maximum $E$ in the early stages, while $\nu$ improves towards negative more gradually. In the early stages of the evolutionary search, the majority of individuals are clustered around zero Poisson's ratio. As the population evolves, the algorithm gradually explores regions with more negative Poisson's ratios, highlighting the complexity of minimizing $\nu$. Notably, Fig. 9(b) shows that each generation's population covers a wide range of property values throughout the evolution process, indicating that diversity is always preserved. We stress that our proposed algorithm is good at balancing exploration and exploitation, demonstrating both rich diversity and stable convergence.

To examine mechanical behaviors explored, we selected six representative metamaterials near the bound of $E$ and $\nu$,as shown in Fig. 9(c) and (d). These unit cells were chosen specifically because they exhibit distinct combinations of $E$ and $\nu$, showcasing extreme mechanical properties across the explored property space. These distinct designs highlight the efficiency of our algorithm to explore the design space.

Additionally, to further elucidate the design space, we conducted a clustering analysis based on genetic similarity and identified five distinct metamaterial families (Fig. 9(e)). Each family exhibits unique geometric features that lead to specific trajectory within the property space. Notably, members of the same family display similar geometric features as the volume fraction increases to enhance stiffness. This family-based classification simplifies the analysis by focusing on shared characteristics within each genotype group, allowing us to identify clear correlations between specific geometric features and mechanical properties. For example, cross-like structures in Family 5 tend to produce higher $\nu$ and higher $E$, while concave designs in Family 3 often lead to more negative $\nu$. Furthermore, as shown in the left panel of Fig. 9(e), each metamaterial family exhibits a trajectory within the property space. This analysis offers the advantage of selecting metamaterials with varying mechanical properties from the same family. Such advantage is critical for resolving incompatible interfaces between neighboring unit cells when assembling heterogeneous microstructures.

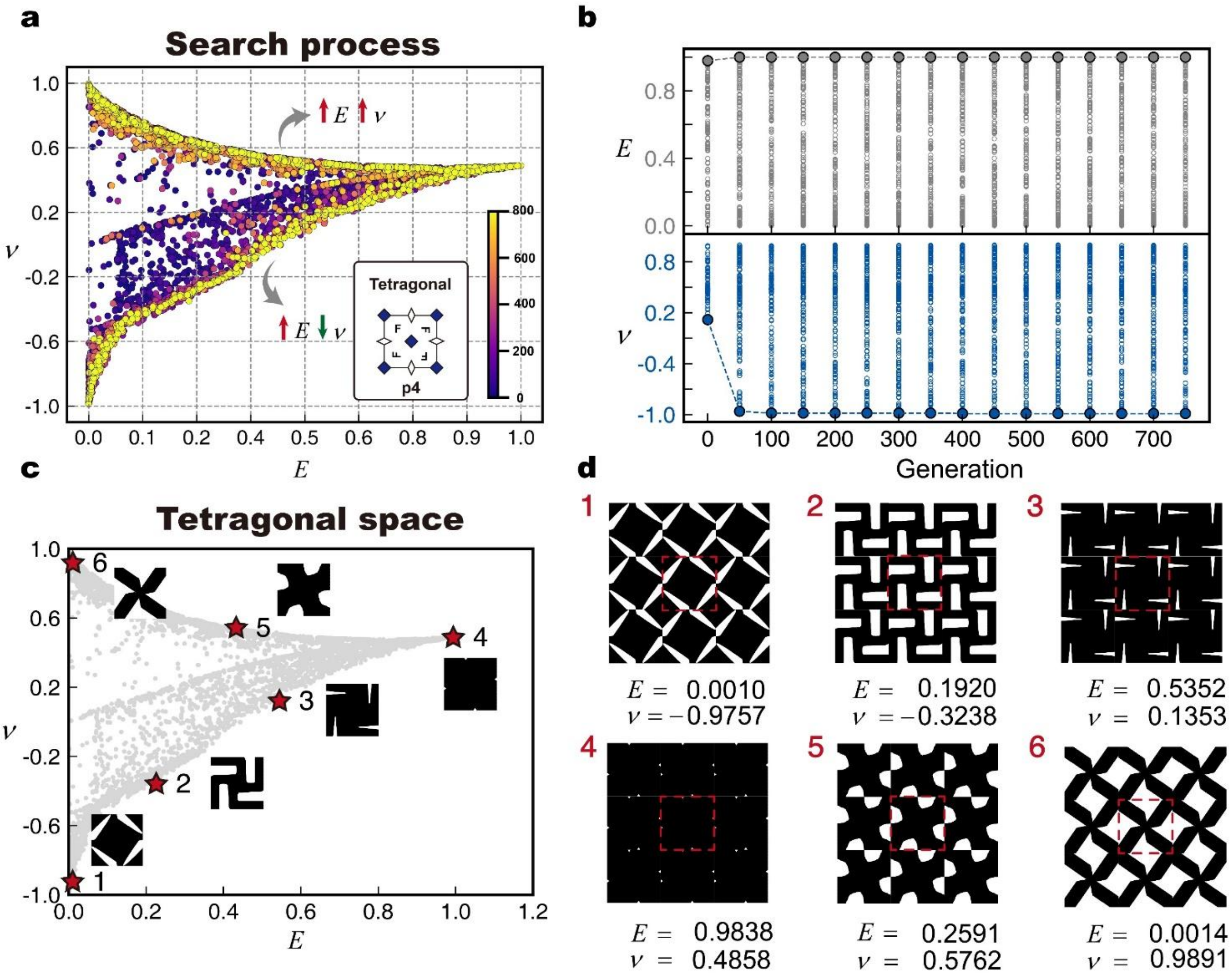

**Fig. 10** Metamaterial designs with the second kind of tetragonal symmetry (p4). (a) The search process and search directions as shown in the property space of Young's modulus $E$ and Poisson's ratio $\nu$. (b) Distribution of individual properties along the evolution. (c) A representative selection of designs on the bond of properties. (d) Enlarged views of the selected designs.

Fig. 10 illustrates our exploration of the design space with rotational symmetry (plane group: p4). Fig. 10(a) shows a uniform distribution along the Pareto front. In contrast to the isotropic designs, the Poisson's ratio of the rotationally symmetric unit cells minimizes rather rapidly, achieving significant negative values within approximately 50 generations (Fig. 10(b)). This result suggests that the inherent rotational symmetry facilitates the emergence of auxetic behavior. In Fig. 10(c) and Fig. 10(d), we select several typical metamaterial designs from the property bound.

### 3.3 Orthotropic Designs

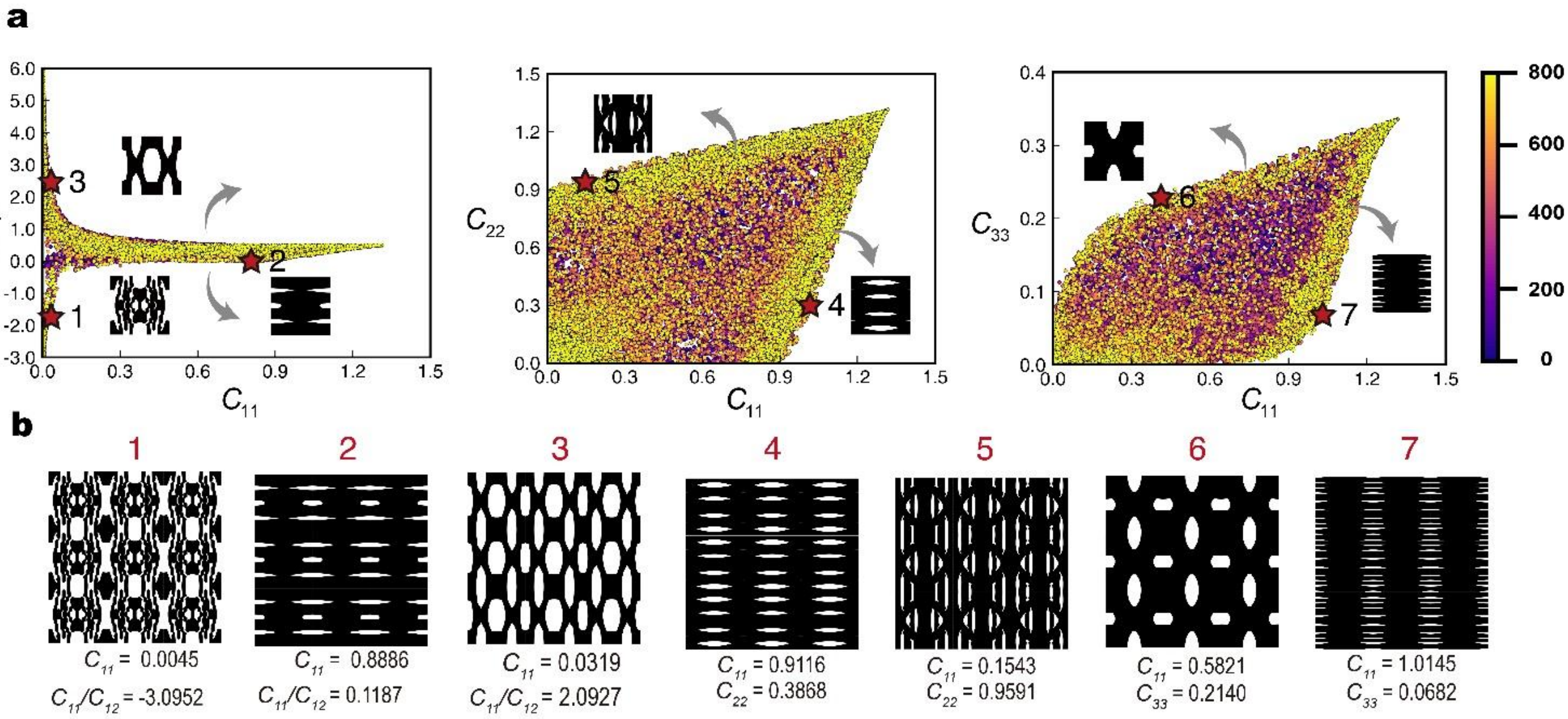

**Fig. 11** Metamaterial designs with the orthogonal symmetry (p2mm). (a) The search process and search directions as shown in the property spaces of different elastic moduli. (b) A representative selection of designs on the bond of properties.

Under orthotropic conditions, we explore the trade-offs between different pairs of elastic moduli under two symmetry planes, as illustrated in Fig. 11. The three pairs examined are $C_{11}/C_{12}$ vs. $C_{11}$, $C_{22}$ vs. $C_{11}$, and $C_{33}$ vs. $C_{11}$, where $\boldsymbol{C}$ denotes the homogenized elasticity tensor. After 800 iterations, we observed that the outcomes from the three pairs are consistent with the results reported in Ref. [5]. This not only validates the effectiveness of our approach but also underscores its capacity for more efficient control in navigating the search space. Unlike the topology optimization driven methods [5], which yields sparse points on the Pareto front of competing properties, our method discovers designs with properties uniformly and densely located on the Pareto front.

### 3.4 Fully Anisotropic Designs

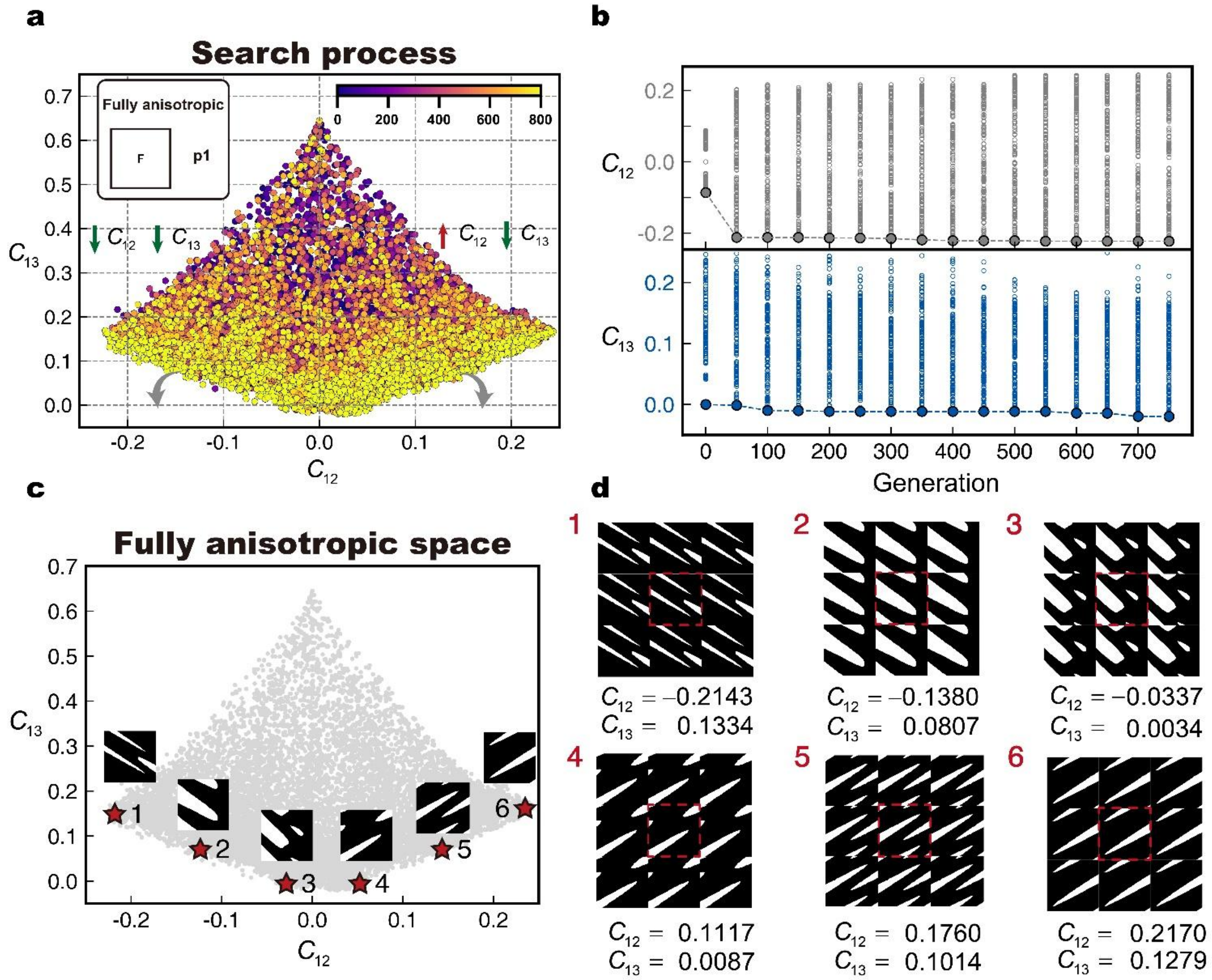

**Fig. 12** Metamaterial designs with fully anisotropic symmetry (p1). (a) The search process and search directions as shown in the property space of $C_{12}$ and $C_{13}$. (b) Distribution of individual properties along the evolution. (c) A representative selection of designs on the bond of properties. (d) Enlarged views of the selected designs.

In the last example, the evolution process aims to explore the shear-normal coupling behavior of fully anisotropic metamaterials, as shown in Fig. 12. Fig. 12(a) shows the evolution process within the property space defined by $C_{12}$ and $C_{13}$, which represent the coupling between normal and shear stresses in different directions [103]. Fig. 12(b) demonstrate convergence trends of each individual property, with $C_{12}$ and $C_{13}$ achieving stable values around generation 800.

Six representative structures are selected, with their properties shown in Fig. 12(c) and geometries shown in Fig. 12(d). We observe that the material is mostly distributed along diagonals of the unit cell to induce a coupled effect between normal stress and shear stress. These microstructural differences underscore the critical role of design in tuning the normal-shear coupling properties of metamaterials. By systematically varying the band orientation, density, and complexity, it is possible to achieve tailored fully anisotropic responses, enabling the design of materials with specific deformation and load-bearing capabilities suited for engineering applications.

### 3.5 Database construction

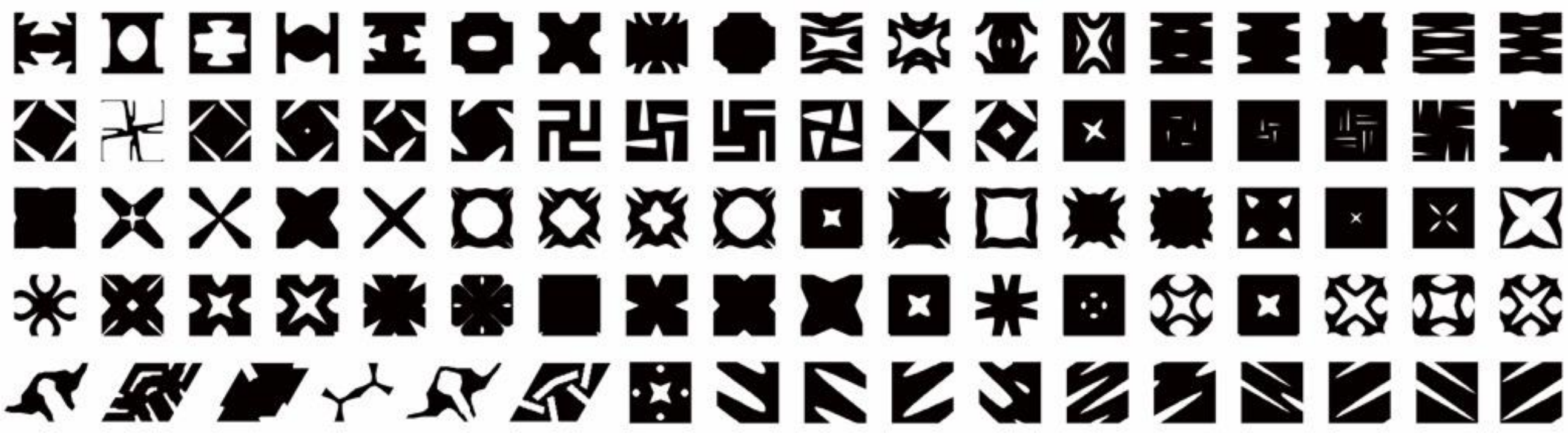

**Fig. 13** Metamaterial database with various kinds of symmetries and properties by lossless encoding of CPPNs, obtained without additional computational cost after the evolutionary search process.

We can collect all data from the evolutions to build an extensive database of metamaterials with various symmetry types (Fig. 13). This approach offers a significant advantage over traditional topology optimization methods, which typically generate only one structure at a time and discard intermediate designs. By collecting all generated designs, we ensure that no computation is wasted, enriching the resource pool for future research and applications. Furthermore, for high-resolution structures, we store only the corresponding implicit CPPN encodings instead of the full explicit representations, with lossless compression. We analyze the number of network hyperparameters in the constructed CPPNs database, and present the results in the following table:

**Table** 1: Statistics of CPPNs nodes and connections

| Statistic | Value |
|---|---|
| Average Node Count | 6.4 |
| Average Connection Count | 7.3 |
| Maximum Node Count | 8 |
| Maximum Connection Count | 16 |

By leveraging the compactness of CPPNs, we drastically reduce memory usage for storing large databases, making the resultant database both comprehensive and scalable.

## 4. Conclusion

A highly scalable, integrated computational framework is proposed for discovery of metamaterial microstructures with diverse and extreme properties. By leveraging the implicit geometry encoding capability of CPPNs, combined with customized neuroevolution algorithm, we evolve the CPPNs without any prior data or gradient information, we obtain a diverse set of designs that are evenly distributed along the Pareto front of competing mechanical properties. We highlight that by combining complex activation functions with spatial regularity and continuous differentiability, the structures generated by CPPNs are both intricate and possess well-defined boundaries. Furthermore, by exploiting the scalability of CPPNs, we can generate metamaterial structures at any resolution using only a minimal number of parameters. Such memory-economic feature allows us to accumulate data from the whole evolution process and build an extensive and diverse database of metamaterials with various symmetry types ensuring that computational resources are fully utilized.

Currently, our work relies on FEM based computational homogenization and focuses solely on two-dimensional linear elasticity. To adapt the proposed framework to three-dimensional cases, we need only to modify the simulation methods to accommodate 3D point clouds without altering the CPPNs and macroevolution algorithm. As the computational cost for 3D metamaterials increases, we plan to incorporate deep learning-based surrogate models alongside the existing Finite Element Method in our framework [104]. This integration will significantly enhance the computational efficiency of the framework, while maintaining its robustness. Moreover, due to the design characteristics of our framework, we can automatically generate the training data needed for the surrogate models, thus eliminating the need for manual preparation of initial training data. We believe the powerful design exploration capability of the proposed framework can be readily extended to other fields by modifying the fitness functions, including thermo-mechanical properties, nonlinear deformations, and other multi-physics behavior.

## CRediT authorship contribution statement

Maohua Yan: Data curation, Formal analysis, Investigation, Methodology, Validation, Visualization, Writing – original draft, Writing – review & editing, Software. Ruicheng Wang: Investigation, Methodology, Validation, Writing – review & editing. Ke Liu: Conceptualization, Funding acquisition, Investigation, Methodology, Project administration, Resources, Supervision, Writing – original draft, Writing – review & editing.

## Declaration of competing interest

The authors declare that they have no known competing financial interests or personal relationships that could have appeared to influence the work reported in this paper.

## Data availability

Data will be made available on request.

## Acknowledgments

This research is supported by the National Natural Science Foundation of China through grant 12372159, the National Key Research and Development Program of China through grant 2022YFB4701900, and the Fundamental Research Funds for the Central Universities, Peking University. The authors thank Chiara Daraio, Josh Bongard, Anya Wallace, Atoosa Parsa, and Sam Kriegman for helpful discussions during the initial development of this research.

## Appendix A. Elastic properties calculation

The elasticity tensor of periodic microstructures are effectively computed through numerical homogenization [35]. We denote the elastic tensor as $\boldsymbol{C}$, and the compliance tensor as $\boldsymbol{S}$, which is the inverse of $\boldsymbol{C}$.

We calculate the effective average Young's modulus and Poisson's ratio based on the components of the compliance tensor. The formulae used for these calculations are given by:

$$E = \frac{1}{2}\left(\frac{1}{S_{11}} + \frac{1}{S_{22}}\right) \tag{17}$$

$$\nu = -\frac{1}{2}\left(\frac{S_{21}}{S_{11}} + \frac{S_{12}}{S_{22}}\right) \tag{18}$$

Here, $S_{ij}$ are the components of the compliance matrix. The formulae provide a convenient way to express the average mechanical properties for materials with orthotropic and anisotropic symmetry. These values serve as a fundamental measure for comparing different metamaterial designs across the design space and are critical for identifying the optimal balance between their properties trade-off.